\documentclass{article}

\usepackage{url}

\usepackage{amsmath,amsfonts}
\usepackage{algorithmic}
\usepackage{array}
\usepackage[caption=false,font=normalsize,labelfont=sf,textfont=sf]{subfig}
\usepackage{textcomp}
\usepackage{stfloats}
\usepackage{verbatim}
\usepackage{graphicx}
\usepackage{multirow}
\def\BibTeX{{\rm B\kern-.05em{\sc i\kern-.025em b}\kern-.08em
    T\kern-.1667em\lower.7ex\hbox{E}\kern-.125emX}}
\usepackage{balance}

\usepackage{booktabs}
\usepackage[ruled,vlined,linesnumbered]{algorithm2e}

\author{Soheil Gharatappeh\thanks{{\tt\small soheil.gharatappeh@maine.edu}, School of Computing and Information Science, University of Maine, Orono, ME, United States}, 
    \and Salimeh Yasaei Sekeh\thanks{Computer Science Department, San Diego State University, San Diego, CA, United States}
    \and Vikas Dhiman\thanks{Department of Electrical and Computer Engineering, University of Maine, Orono, ME, United States}%
    \thanks{This material is partially based upon work supported by the National Science Foundation under Grant No 2218063}
}

\title{Report: Weather-Aware Object Detection Transformer for Domain Adaptation}

\begin{document}
\maketitle

\begin{abstract}
RT-DETRs have shown strong performance across various computer vision tasks but are known to degrade under challenging weather conditions such as fog. In this work, we investigate three novel approaches to enhance RT-DETR robustness in foggy environments: (1) Domain Adaptation via Perceptual Loss, which distills domain-invariant features from a teacher network to a student using perceptual supervision; (2) Weather Adaptive Attention, which augments the attention mechanism with fog-sensitive scaling by introducing an auxiliary foggy image stream; and (3) Weather Fusion Encoder, which integrates a dual-stream encoder architecture that fuses clear and foggy image features via multi-head self and cross-attention. Despite the architectural innovations, none of the proposed methods consistently outperform the baseline RT-DETR. We analyze the limitations and potential causes, offering insights for future research in weather-aware object detection.
\end{abstract}

\section{Introduction}
Object detection algorithms have witnessed remarkable advancements in recent years, evolving from computationally expensive two-stage approaches \cite{ren-2015-faster-r-cnn, girshick2015fast} to highly efficient single-stage architectures \cite{redmon2016look, liu-2016-ssd}.
Early methods such as Faster R-CNN \cite{ren-2015-faster-r-cnn} demonstrated strong performance but suffered from high latency due to their region proposal mechanisms.
The advent of single-stage detectors like YOLO and SSD revolutionized real-time object detection by eliminating the need for proposal generation, relying instead on direct regression and classification.
However, these architectures heavily incorporate hand-crafted components such as anchor boxes and non-maximum suppression, which impose constraints on their adaptability to diverse environments.

More recently, transformer-based architectures have emerged as a powerful alternative to CNN-based networks \cite{dosovitskiy_image_2021}, demonstrating superior performance on various computer vision benchmarks.
Models such as DETR \cite{carion_end--end_2020} and its variants \cite{zhang2022dino, liu2024grounding} leverage the self-attention mechanism to model long-range dependencies and learn object representations with greater flexibility.
Unlike CNNs, transformers do not rely on hand-crafted design choices, making them more adaptable to different visual contexts~\cite{dosovitskiy_image_2021}.
However, despite these advantages, transformer-based detectors remain highly sensitive to image degradation caused by adverse weather conditions, such as fog \cite{yu_mttrans_2022}. In real-world scenarios, fog significantly decreases image quality by reducing contrast, obscuring fine details, and introducing scattering effects, which severely impact the feature extraction capability of these models.

Unsupervised Domain Adaptation (UDA) has been extensively studied to address domain shifts arising from variations between a labeled source domain and an unlabeled target domain \cite{chen2018domain, li2022cross, zhang_da-detr_2023}. UDA techniques typically focus on aligning feature distributions between domains to facilitate knowledge transfer. This is often achieved through adversarial learning \cite{he2019multi}, self-supervised objectives~\cite{ZHOU2023103649}, or statistical feature alignment \cite{zhao_task-specific_2022}. However, these methods primarily emphasize the alignment of cross-domain features, which can inadvertently lead to the loss of discriminatory information. Specifically, in adverse weather conditions, UDA approaches may struggle to disentangle domain-specific artifacts (e.g., fog-induced distortions) from essential object features, leading to suboptimal detection performance.

To overcome these limitations, we explore alternative strategies that emphasize feature robustness without relying solely on domain alignment. Specifically, we propose integrating perceptual loss into transformer-based detectors, specifically RT-DETRs, to enhance robustness. Originally introduced for tasks like image synthesis and style transfer \cite{johnson-2016-percep-losses}, perceptual loss preserves high-level semantic features by minimizing discrepancies in deep feature space rather than at the pixel level. By integrating this loss into a knowledge distillation based  pipeline \cite{hu-2022-teach-studen}, we aim to encourage the model to learn domain-invariant yet discriminative representations, helping to retain critical object features despite fog-induced degradation.

In addition to feature-level alignment, we explore a weather-adaptive attention mechanism that enhances the model’s ability to modulate attention based on the context of the fog. This approach introduces an auxiliary stream that processes a foggy version of the input image, producing scalar weights that adjust the attention computation in the transformer encoder. This dynamic modulation allows the model to better emphasize relevant features even when visibility is reduced.

Finally, we propose a weather fusion encoder that operates on both clear and foggy versions of the input image in parallel. Each stream undergoes self-attention refinement, and their features are subsequently fused through a cross-attention module. This design enables the model to integrate complementary cues from both domains, promoting more robust and adaptive object detection in visually challenging foggy scenes.

The main contributions of this paper are:
\begin{itemize}
    \item We proposed domain adaptation through perceptual loss and knowledge transfer through a teacher network.
    \item We proposed FogAwareAttention, an attention mechanism that adapts RT-DETR encoder values to foggy weather by incorporating the foggy image pipeline.
    \item We proposed a Weather Fusion Encoder where a cross-attention mechanism is used to fuse the clear image and foggy image embeddings.
\end{itemize}

The remainder of this paper is structured as follows. Section 2 reviews related work on object detection, domain adaptation, perceptual loss, and RT-DETRs. Section 3 introduces our proposed method and details the integration of three proposed methods. Section 4 presents experimental results, including comparisons with existing methods under foggy conditions. Finally, Section 5 concludes the paper and discusses potential future research directions.

\section{Related Work}

\subsection{Object Detection Using Transformers}
Deep convolutional neural networks (CNNs) have become the standard approach for object detection, with two-stage methods such as Faster R-CNN \cite{ren-2015-faster-r-cnn} and one-stage approaches like YOLO and SSD \cite{redmon2016look, liu-2016-ssd} demonstrating strong performance on large-scale benchmarks.
Despite their success, these methods rely heavily on handcrafted components such as non-maximum suppression (NMS), Intersection over Union (IoU) thresholds, and anchor box design.
As a result, they often require extensive fine-tuning and cannot be trained in an end-to-end fashion.

Transformers, originally designed for sequence modeling in natural language processing, have recently been adapted for vision tasks.
Vision Transformers (ViTs) showed promising results but were initially limited by their high data requirements, often underperforming CNNs when trained on smaller datasets.
DEtection TRansformers (DETRs) addressed this by combining a CNN backbone with a transformer-based encoder-decoder structure, leveraging multi-head self-attention to model global context effectively \cite{carion_end--end_2020}.
This made DETRs the first end-to-end object detection framework based on transformers.

Real-Time DETRs (RT-DETRs) further improved efficiency and accuracy by introducing a multi-scale feature pyramid, an Efficient Hybrid Encoder, and a Cross-Scale Feature Fusion (CSFF) block \cite{zhao_detrs_2024}.
These components enhanced feature representation at multiple resolutions and enabled the detection of small objects—an area where traditional transformers often struggled.
RT-DETR’s streamlined design allows for real-time inference while preserving the benefits of transformer-based modeling.

\subsection{Domain Adaptation for Object Detection}
Unsupervised Domain Adaptation (UDA) has been extensively studied to address performance degradation caused by distribution shifts between a labeled source domain and an unlabeled target domain. Existing UDA methods for object detection are primarily categorized into two levels: domain-level alignment and category-level alignment.

Domain-level UDA focuses on reducing the overall distribution discrepancy between the source and target domains. This is typically achieved by aligning global feature distributions across domains using statistical measures such as Maximum Mean Discrepancy \cite{zhang2013domain, tzeng2017adversarial}] and Correlation Alignment \cite{sun2016deep}. These methods attempt to map source and target features into a shared feature space by minimizing distributional divergence, either at the image level or the feature level. Adversarial training strategies have also been employed in this context to learn domain-invariant representations by pitting a feature extractor against a domain discriminator \cite{tsai2017adversarial}.

Category-level UDA offers a more fine-grained approach to domain adaptation by aligning class-conditional distributions between the source and target domains.
Techniques such as \cite{saito2017asymmetric}, \cite{cai2019exploringobjectrelationmean}, and \cite{hoyer2022daformerimprovingnetworkarchitectures} employ adversarial training objectives, where dual classifiers are optimized to maximize prediction discrepancies on target samples.
This, in turn, drives the feature extractor to learn more discriminative and class-aware representations.
While this strategy effectively reduces intra-class domain discrepancies, it remains sensitive to noisy labels and class imbalance—challenges that are especially pronounced in dense prediction tasks like object detection.

Recently, category-level UDA has also been extended to transformer-based architectures.
For example, CDTrans \cite{xu_cdtrans_2022} demonstrates that integrating triple-branch self-attention and cross-attention mechanisms can enhance feature alignment quality within the transformer framework.
This adaptation not only improves domain transfer performance but also highlights the potential of leveraging transformer-specific structures for robust domain adaptation in vision tasks.
However, despite these advancements, UDA methods often struggle to disentangle domain-specific artifacts—such as weather-induced degradations—from essential semantic features, leading to suboptimal generalization in real-world scenarios.

\subsection{Knowledge Distillation}
To enhance robustness in domain adaptation, recent approaches have increasingly turned to knowledge distillation as a complementary strategy. Originally introduced as a model compression technique, knowledge distillation transfers the knowledge of a large, high-performing teacher network to a smaller or less robust student network by encouraging the student to replicate the teacher’s output distributions or intermediate representations \cite{hinton2015distilling}. This paradigm has proven effective across a wide range of tasks—including classification, object detection, and domain adaptation—particularly in scenarios involving limited supervision or noisy input conditions.

In a typical teacher-student framework, the teacher network—pretrained on clean, high-quality source domain data—generates soft targets or feature embeddings that guide the student network, which learns on more challenging or degraded data \cite{romero2014fitnets, chen2021distilling}. Recent efforts have extended this concept to transformer-based models. For example, Mean Teacher strategies have been applied to transformers to align multi-level features through pseudo-labeling \cite{yu_mttrans_2022}. ONDA-DETR \cite{suzuki_onda-detr_2023} further leverages this idea by generating pseudo-labels for unlabeled target-domain samples using a recall-aware labeling strategy combined with quality-aware training to improve detection performance under domain shift.
However, a persistent challenge in pseudo-label–based domain adaptation remains: the propagation of incorrect pseudo-labels, which can misguide the student network and hinder generalization if not adequately filtered or weighted.

Inspired by these advances, our work adopts a similar teacher-student structure to facilitate perceptual alignment between clear and foggy domains. This setup enables the student model to learn domain-invariant, yet semantically rich representations, ultimately leading to more robust object detection in adverse weather conditions.

\subsection{Perceptual Loss}
Perceptual loss, originally introduced in the context of image generation and style transfer \cite{johnson-2016-percep-losses}, measures the discrepancy between high-level features extracted from a pre-trained deep neural network rather than relying solely on low-level pixel-wise differences.
By comparing deep feature representations, perceptual loss captures semantic similarities between images, which makes it particularly effective in tasks where visual distortions (e.g., noise, blur, or fog) alter pixel values but not the underlying object structure.
In recent years, perceptual loss has been integrated into object detection and domain adaptation pipelines to encourage the preservation of semantic content across domains \cite{hong2024tackling, cheng2021dual}.
This is especially beneficial in adverse weather conditions where direct feature alignment may fail due to degraded visibility.
Our approach leverages perceptual loss to guide a student network trained on foggy data to mimic the internal representations of a teacher network exposed to clear data, thereby facilitating more robust and discriminative feature learning across domains.

Despite the progress, these methods often struggle under adverse weather conditions such as fog, where domain-specific artifacts distort visual features and mislead the alignment process. Our work addresses these limitations by proposing transformer-based architectures that incorporate perceptual and attention-based mechanisms for more robust adaptation. By explicitly preserving high-level semantics and leveraging cross-domain attention fusion, our method improves adaptation robustness, especially under challenging visual distortions.

Dehazing algorithms aim to restore clear images from foggy images \cite{kaiming_2011_single_image, cai2016dehazenet}. While dehazing can improve the performance of object detection models, it can also introduce artifacts and may not always be effective in extreme fog conditions.

\section{Background}

Transformer-based architectures have achieved remarkable success in a variety of vision and language tasks, largely due to their attention mechanisms. In the context of vision transformers such as RT-DETR, self-attention plays a critical role in modeling long-range dependencies and context across the spatial features of an image.

\subsection{Self-Attention}

Given an input sequence of features $\mathbf{X} \in \mathbb{R}^{n \times d}$, where $n$ is the number of tokens and $d$ is the feature dimension, self-attention is computed as follows:

\begin{equation}
\mathbf{Q} = \mathbf{X}\mathbf{W}_Q, \quad
\mathbf{K} = \mathbf{X}\mathbf{W}_K, \quad
\mathbf{V} = \mathbf{X}\mathbf{W}_V
\end{equation}

where $\mathbf{W}_Q, \mathbf{W}_K, \mathbf{W}_V \in \mathbb{R}_{d \times d_k}$ are learnable weight matrices that project the input into query, key, and value representations, respectively. The self-attention output is then computed as:

\begin{equation}
\text{SA}(\mathbf{Q}, \mathbf{K}, \mathbf{V}) = \text{softmax}\left(\frac{\mathbf{Q}\mathbf{K}^\top}{\sqrt{d_k}}\right)\mathbf{V}
\end{equation}
The dot product between queries and keys determines how much focus each token places on the others, and the result is used to weigh the value vectors accordingly. To capture diverse contextual relationships from different representation subspaces, transformer models use multi-head self-attention. This involves projecting the input into multiple independent sets of queries, keys, and values. For $h$ attention heads, the computation is:

\begin{equation}
\text{MultiHead}(\mathbf{Q}, \mathbf{K}, \mathbf{V}) = \text{Concat}(\text{head}_1, \dots, \text{head}_h)\mathbf{W}_O
\end{equation}

\begin{equation}
\text{where} \quad \text{head}_i = \text{Attention}(\mathbf{Q}_i, \mathbf{K}_i, \mathbf{V}_i), \quad \mathbf{Q}_i = \mathbf{X}\mathbf{W}_{Q_i}, \ \mathbf{K}_i = \mathbf{X}\mathbf{W}_{K_i}, \ \mathbf{V}_i = \mathbf{X}\mathbf{W}_{V_i}
\end{equation}

Each head attends to the input from a different subspace, and their outputs are concatenated and projected using $\mathbf{W}^O \in \mathbb{R}^{hd_k \times d}$. This formulation enables the model to attend to information at multiple positions and granularities simultaneously.

\subsection{Perceptual Loss}
Perceptual loss is designed to maintain high-level feature consistency between clean and foggy images. Instead of relying purely on pixel-wise loss functions such as mean squared error (MSE), which may fail to capture structural information, perceptual loss compares feature activations extracted from a pre-trained network. We employ a CNN to extract multi-scale feature representations $ \phi(I) $ at different layers.

The perceptual loss is defined as:
\begin{equation} \label{eq:perc_loss}
\mathcal{L}_{perc} = \sum_{l \in P} \lambda_l | f^{(l)}(I_s) - f^{(l)}(I_t) |^2
\end{equation}
where $ P $ is the set of layers in the network $ \mathcal{F} $ included in perceptual loss computation, and $ \lambda_l $ are weighting coefficients that control the contribution of each layer.

\subsection{Fog}
Foggy weather conditions introduce significant challenges to object detection by reducing visibility and altering the statistical properties of images. The degradation caused by fog can be modeled using the atmospheric scattering model, which describes the observed intensity of a scene point in a foggy image $ I_t(x) $ as:

\begin{equation}\label{eq:fog}
I_t(x) = I_s(x) e^{-\beta d(x)} + A(1 - e^{-\beta d(x)})
\end{equation}
where $ I_s(x) $ represents the scene radiance (the true underlying image), $ A $ is the atmospheric light, $ \beta $ is the scattering coefficient that determines the density of fog, and $ d(x) $ is the scene depth.
This equation captures how light is attenuated as it travels through fog, leading to reduced contrast and detail loss in distant objects.
Traditional object detectors, including transformer-based RT-DETR models, struggle to maintain performance under these conditions due to the uncertainty in uncertainties in the estimation of key parameters such as atmospheric light and scattering coefficient, as well as epistemic errors in the model's representation of degraded visual conditions.

Given an input image $ I_t $ affected by fog, our goal is to train a detector $ \mathcal{F} $ that is robust to these conditions.
Formally, let $ I_s $ be a clean image from the source domain and $ I_t $ be a foggy image from the target domain.
Also, let $ f^{(l)}(I) $ be the feature map that is produced by the CNN backbone  network $G(.)$, used for feature extraction, when fed with image $ I $. 


\section{Methodology}
In this section, we present three approaches designed to improve the robustness of RT-DETR under foggy conditions. Each method targets a different aspect of the detection pipeline to address the domain shift introduced by adverse weather. First, we propose a domain adaptation strategy based on perceptual loss to enforce feature-level consistency across domains. Second, we introduce a weather adaptive attention mechanism that dynamically modulates the transformer’s attention based on fog context. Finally, we develop a weather fusion encoder that integrates complementary features from both clear and foggy image streams. The following subsections detail the design and implementation of each method.

\subsection{Perceptual Loss RT-DETR}
\begin{figure}
  \centering
 \includegraphics[scale=.4]{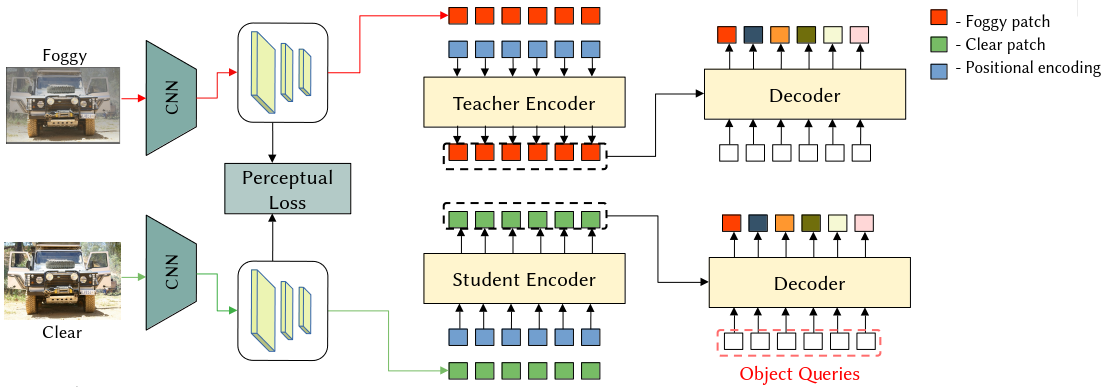} 
  \caption{RT-DETR Perceptual Loss}
  \label{fig:rtdetr_perc}
\end{figure}

Perceptual Loss RT-DETR (PL-RT-DETR) incorporates perceptual loss (Equation~\eqref{eq:perc_loss}) into the RT-DETR framework to facilitate knowledge transfer from a source domain (clear images) to a target domain (foggy images), as illustrated in Figure~\ref{fig:rtdetr_perc}. This method is inspired by the concept that perceptual similarity in deep feature space better preserves semantic content than pixel-wise losses, particularly under domain shifts caused by visual distortions like fog.

In this setup, a teacher network pretrained on clear-weather images processes a clear input image and produces high-level patch embeddings that capture clean, domain-consistent features.
Simultaneously, a student network, fed with the foggy counterpart of the same image, generates fog-domain embeddings.
The perceptual loss then measures the discrepancy between the teacher and student embeddings in the feature space of the detection backbone, encouraging the student to learn representations that remain semantically consistent with those from the clear domain. The Total loss is defined as

\begin{equation}
  \label{eq:1}
  \mathcal{L} = \mathcal{L}_{\text{obj}} + \mathcal{L}_{\text{perc}}
\end{equation}
where $\mathcal{L}_{\text{obj}}$ is the original detection loss defined in RT-DETR.

This alignment enables the student network to retain meaningful object-level information even in the presence of fog, enhancing detection performance under adverse conditions without requiring explicit supervision on the target domain.

\subsection{Weather Adaptive Attention for RT-DETR}
\begin{figure}
  \centering
 \includegraphics[scale=.4]{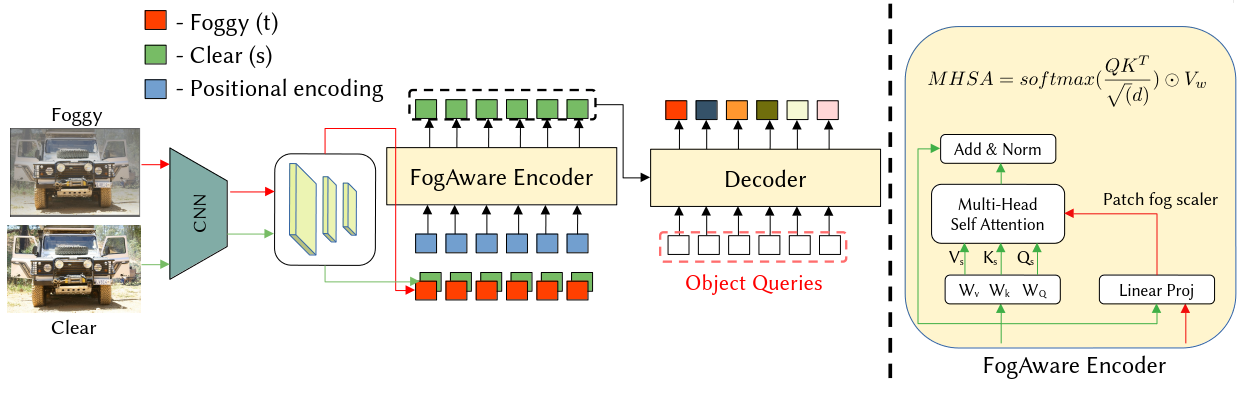}
  \caption{Fog-Aware Attention}
  \label{fig:fogaware_attention}
\end{figure}

Weather Adaptive Attention (WAA-RT-DETR) modifies the standard RT-DETR attention mechanism by incorporating a fog data stream into the network \ref{fig:fogaware_attention}. Fog, as shown in equation \eqref{eq:fog} is dependent on depth, as such the depth channel can be used as a proxy to how fog can impact the image. In Fog-Aware attention, the fog density map is used to adjust the attention weights, allowing the model to focus on relevant features even in the presence of fog.

Let $I_t$ be the depth image, and $W_t$ be a linear projection used to project the output embedding $z(I_t)$ produced by adding flattened backbone features to the positional encoding. Weather scalar $V_w$ is defined as
\begin{equation}
  \label{eq:weather_scalar}
  V_w = W_t z(I_t)
\end{equation}

Let's define fog attention as
\begin{equation}
  \label{eq:fogaware}
\text{SA}_{fog}(Q, K, V) = \text{softmax}\left(\frac{QK^T}{\sqrt{d_k}} \odot V_w\right)V
\end{equation}
where $\odot$ represents element-wise multiplication. This modification is applied within the attention modules of RT-DETR to effectively adapt the model to foggy conditions.

The intuition behind this modification is that regions with high fog density should have lower attention weights, allowing the model to focus on regions with less fog, improving object detection in foggy scenarios.

\begin{figure}
  \centering
  \includegraphics[scale=0.4]{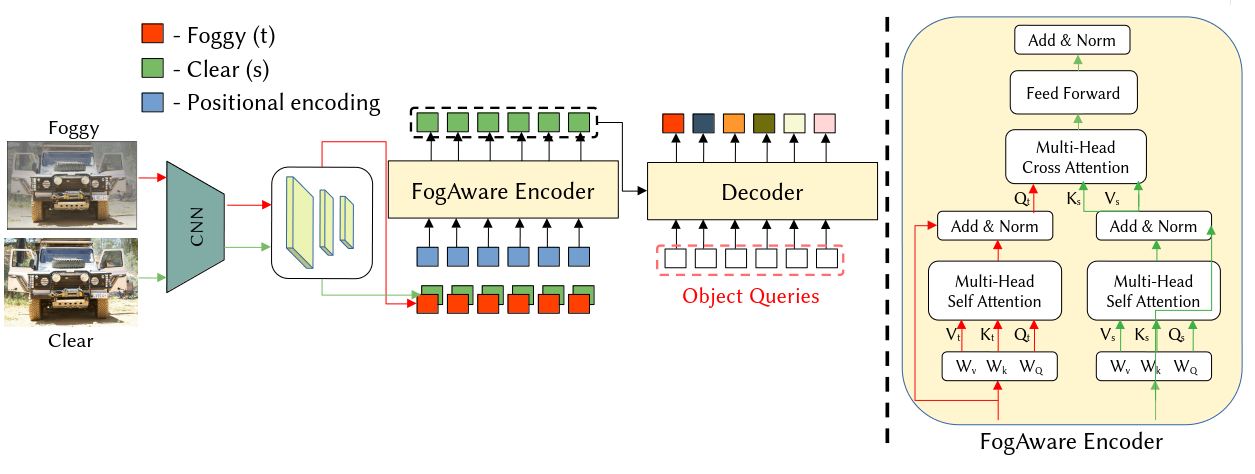}
  \caption[Weather Adaptive Attention]{Dual Setl-Attention + Cross-Attention Fusion}
  \label{fig:daul_wa}
\end{figure}

\subsection{Weather Fusion Encoder}
To effectively leverage complementary information from both clear and foggy images, we propose a Weather Fusion Encoder (WFE-RT-DETR) composed of parallel self-attention modules followed by a cross-attention-based fusion mechanism.

Let $\mathbf{X}_{\text{img}} \in \mathbb{R}^{n \times d}$ and $\mathbf{X}_{\text{fog}} \in \mathbb{R}^{n \times d}$ denote the input embeddings extracted from the clear and foggy images, respectively. We first apply self-attention independently to each stream:

\begin{equation}
\mathbf{E}_{\text{img}} = \text{SA}(\mathbf{X}_{\text{img}}, \mathbf{X}_{\text{img}}, \mathbf{X}_{\text{img}}), \quad
\mathbf{E}_{\text{fog}} = \text{SA}(\mathbf{X}_{\text{fog}}, \mathbf{X}_{\text{fog}}, \mathbf{X}_{\text{fog}})
\end{equation}

To enable interaction between the two representations, we apply a cross-attention operation where the contextualized clear features query the foggy stream:

\begin{equation}
\mathbf{E}_{\text{cross}} = \text{CA}(\mathbf{E}_{\text{img}}, \mathbf{E}_{\text{fog}}, \mathbf{E}_{\text{fog}})
\end{equation}
Lastly, the fused representation is then passed through a residual and normalization layer and is used as the input for the decoder layer.




This design allows the model to integrate domain-specific cues from the foggy image while preserving structural clarity from the clean image, enhancing the robustness of the learned features under foggy conditions.

\section{Experiments}
\subsection{Datasets}
We evaluate our proposed methods on both synthetic and real-world foggy datasets to assess robustness under adverse weather conditions.

For the synthetic dataset, a synthetic fog is generated and added to the clear-weather images using the atmospheric scattering model (Equation~\eqref{eq:fog}).

For real-world evaluation, we use the RTTS dataset, which comprises approximately 4,000 images captured in actual foggy conditions. RTTS includes annotations for five object categories: \textit{bicycle}, \textit{bus}, \textit{car}, \textit{motorbike}, and \textit{person}. To ensure consistency during evaluation, all other datasets used in our experiments are filtered to include only these five categories.

In addition to RTTS, we incorporate the Pascal VOC dataset, which contains approximately 16,000 training images and 4,000 validation images spanning various indoor and outdoor scenes. While Pascal VOC includes 20 object categories, we filter the dataset to retain only the five categories present in RTTS, allowing for aligned training and evaluation across domains.

\subsection{Implementation Details}
We implement the proposed Fog-Aware Attention and Weather Fusion Encoder by modifying the standard RT-DETR architecture provided by the Hugging Face Transformers library. For the perceptual loss approach, we extend the Ultralytics implementation of RT-DETR to incorporate a teacher-student framework for domain adaptation.

All models are initialized with publicly available pretrained weights and fine-tuned on the source (clear-weather) datasets. Evaluation is performed on the corresponding target (foggy) datasets to assess generalization and robustness under adverse weather conditions.

\subsection{Experimental Results}
\textbf{Weather Adaptation}: In this experiment, we compare the performance of PL-RT-DETR with the vanilla RT-DETR and YOLO-v8, that is the current baseline for real-time object detection.
We trained the Teacher network for 100 epochs with a mixture of foggy and clear data, with the fog severity being chosen randomly on the fly, and then transfer the knowledge from the Teacher to the Student network by training the student network for 100 epochs.

\begin{table*}[ht]
\centering
\caption{Comparison of per-category and overall mAP for various models trained on VOC and evaluated on VOC, RTTS, and synthetic fog datasets.}
\label{tab:voc_training}
\small
\begin{tabular}{llcccccc}
\toprule
\textbf{Model} & \textbf{Eval. Set} & \textbf{Car} & \textbf{Bus} & \textbf{Bicycle} & \textbf{M.bike} & \textbf{Person} & \textbf{mAP} \\
\midrule
  \multirow{5}{*}{YOLOv8}
    & VOC       & 0.922 & 0.881 & 0.907 & 0.905 & 0.898 & 0.903 \\
    & RTTS      & 0.418 & 0.136 & 0.319 & 0.231 & 0.471 & 0.315 \\
    & Low Fog   & 0.920 & 0.867 & 0.907 & 0.897 & 0.891 & 0.896 \\
    & Mid Fog   & 0.918 & 0.863 & 0.909 & 0.893 & 0.889 & 0.895 \\
    & High Fog  & 0.913 & 0.856 & 0.909 & 0.887 & 0.887 & 0.890 \\

\midrule
\multirow{5}{*}{RT-DETR} 
    & VOC       & 0.919 & 0.911 & 0.932 & 0.886 & 0.899 & 0.909 \\
    & RTTS      & 0.534 & 0.167 & 0.457 & 0.280 & 0.577 & 0.403 \\
    & High Fog  & 0.764 & 0.750 & 0.885 & 0.825 & 0.775 & 0.800 \\
    & Mid Fog   & 0.763 & 0.749 & 0.886 & 0.825 & 0.775 & 0.800 \\
    & Low Fog   & 0.763 & 0.749 & 0.885 & 0.825 & 0.775 & 0.800 \\

\midrule
\multirow{5}{*}{PL-RT-DETR} 
    & VOC       & 0.919 & 0.911 & 0.932 & 0.886 & 0.899 & \textbf{0.909} \\
    & RTTS      & 0.559 & 0.206 & 0.435 & 0.284 & 0.627 & \textbf{0.422} \\
    & High Fog  & 0.852 & 0.843 & 0.927 & 0.873 & 0.859 & 0.871 \\
    & Mid Fog   & 0.852 & 0.847 & 0.928 & 0.875 & 0.860 & 0.872 \\
    & Low Fog   & 0.850 & 0.848 & 0.925 & 0.873 & 0.859 & 0.871 \\
\bottomrule
\end{tabular}
\end{table*}
Table~\ref{tab:voc_training} presents a comprehensive comparison of three object detection models—YOLOv8, RT-DETR, and PL-RT-DETR—trained on the VOC dataset and evaluated across multiple domains, including VOC, RTTS (real-world adverse weather), and synthetic fog conditions (low, mid, and high fog levels). The performance is reported in terms of class-wise average precision and overall mean Average Precision at IoU threshold 0.5 (mAP@50).

On the clean VOC dataset, all models perform similarly well, with mAP values above 0.90. Notably, PL-RT-DETR and RT-DETR both achieve the highest mAP of 0.909, slightly outperforming YOLOv8 (0.903), demonstrating that transformer-based detectors can match or exceed convolutional baselines under favorable conditions.

However, under domain shift to RTTS, significant performance degradation is observed for all models, highlighting the challenges of generalizing to adverse weather. PL-RT-DETR outperforms both RT-DETR and YOLOv8 with a mAP of 0.422, marking a substantial improvement over RT-DETR (0.403) and YOLOv8 (0.315). This suggests that PL-RT-DETR benefits from enhanced robustness to real-world degradation factors, possibly due to its learned representations being less sensitive to domain shift.

Under synthetic fog scenarios, PL-RT-DETR continues to demonstrate superior performance across all fog levels, achieving consistent mAP values of approximately 0.871–0.872.

Overall, PL-RT-DETR achieves the highest robustness and cross-domain generalization, outperforming baselines in challenging environments without sacrificing performance on the source domain. These findings validate the effectiveness of our proposed model for weather-adaptive detection tasks.

\section{Conclusion}
In this work, we explored three extensions to the RT-DETR architecture aimed at improving object detection performance under foggy conditions.
Through extensive experiments and evaluation across clear and degraded domains, we compared each proposed method against the RT-DETR baseline.
Among the three variants, PL-RT-DETR demonstrated a slight but consistent improvement, validating the effectiveness of perceptual loss for domain adaptation.
The WFE-RT-DETR variant performed on par with the baseline, indicating that while the added module may not hinder performance, further tuning may be needed to realize its full potential.
On the other hand, WAA-RT-DETR failed to converge during training.
We attribute this to the multiplicative nature of its attention mechanism, which likely introduces instability and requires additional refinement to become practically effective.
Overall, our findings provide insight into the challenges and trade-offs of adapting transformer-based detectors to adverse weather conditions and highlight promising directions for future improvement.
We have proposed Fog-Aware Attention, a novel attention mechanism that adapts RT-DETR to foggy weather conditions.
Our method incorporates a fog density estimation module and dynamically adjusts the attention weights based on the estimated fog density.
Experimental results demonstrate the effectiveness of our approach in improving object detection performance in foggy environments.

Future work will focus on extending Fog-Aware Attention to other adverse weather conditions and exploring more efficient fog density estimation modules, as well as testing on more complex real-world datasets.

\bibliographystyle{plain}
\bibliography{rtdetr-main}


\end{document}